\documentclass{article} 
\usepackage{iclr2027_conference,times}

\usepackage{amsmath,amsfonts,bm}

\def\eqref#1{equation~\ref{#1}}

\def\1{\bm{1}}

\DeclareMathAlphabet{\mathsfit}{\encodingdefault}{\sfdefault}{m}{sl}
\SetMathAlphabet{\mathsfit}{bold}{\encodingdefault}{\sfdefault}{bx}{n}

\usepackage{booktabs}
\usepackage{multirow}
\usepackage{placeins}
\usepackage{hyperref}
\usepackage{url}
\usepackage{enumitem}
\usepackage{amsmath} 
\usepackage[most]{tcolorbox}
\newtcolorbox{agentbox}{
    colback=blue!3,
    colframe=blue!45!black,
    boxrule=0.5pt,
    arc=2pt,
    left=6pt,
    right=6pt,
    top=6pt,
    bottom=6pt,
    breakable
}

\newtcolorbox{humanbox}{
    colback=green!3,
    colframe=green!40!black,
    boxrule=0.5pt,
    arc=2pt,
    left=6pt,
    right=6pt,
    top=6pt,
    bottom=6pt,
    breakable
}
\usepackage[utf8]{inputenc} 
\usepackage[T1]{fontenc}    
\usepackage{booktabs}       
\usepackage{amsfonts}       
\usepackage{nicefrac}       
\usepackage{microtype}      
\usepackage{xcolor}         
\usepackage{graphicx}       
\usepackage{float}          
\usepackage{placeins}       

\usepackage{longtable}
\usepackage{booktabs}
\usepackage{array}
\usepackage{ragged2e}

\title{Process Matters more than Output for \\ Distinguishing Humans from Machines}

  \author{%
    Milena Rmus \\
    Proof of Human, PBC \\                                                                                                      
    \texttt{milena.rmus@poh.org} \\
    \And                                                                                                                
    Mathew D. Hardy \\
    Proof of Human, PBC \\                                                                                                      
    \texttt{matt.hardy@poh.org} \\
    \And                                                                                                                
    Thomas L.~Griffiths \\
    Princeton University \\                                                                                                      
    \texttt{tomg@princeton.edu} \\
    \And
    Mayank Agrawal \\                                                                                                   
    Proof of Human, PBC \\
    \texttt{mayank.agrawal@poh.org} \\                                                                                                   
  }

\iclrfinalcopy 
\begin{document}

\maketitle

\begin{abstract}
Reliable human-machine discrimination is becoming increasingly important as Large Language Models and autonomous agents are deployed in online settings. Existing approaches largely evaluate whether a system can produce behavior or responses indistinguishable from those of a human. This approach follows the focus on the \textit{output} of a machine as a criterion for determining whether it can think, as suggested by Alan Turing. Cognitive science provides an alternative approach: considering the \textit{process} by which that behavior is produced. To evaluate whether differences in cognitive mechanisms can reliably distinguish humans from machines, we introduce a process-based framework, the \textit{Process Turing Test}, and evaluate it across a battery of cognitive tasks spanning decision-making, working memory, and planning. These tasks, such as \textit{mental rotation} and \textit{sequence prediction}, yield process-level measures that reveal how behavior is generated, complementing conventional measures of overall task performance. We also include multiple CAPTCHA tasks in the battery to compare the experimental paradigms to challenges deployed in the real-world. Across the battery, process-level features provide substantially stronger discriminative signal than performance metrics alone, reliably distinguishing humans from agents even when task performance is matched (mean process-feature classifier AUC = 0.88). To assess agentic process limitations, we conduct a controlled red-teaming study comparing off-the-shelf frontier agents (Claude Sonnet 4.5, GPT-5, Gemini 2.5 Pro), Centaur (a large language model fine-tuned on 10.7M human decisions), and two task-specific fine-tuning methods applied to models of different sizes and families: \textbf{action-level supervised fine-tuning (A-SFT)} and \textbf{process-level fine-tuning (P-SFT)}, which directly optimizes process features. We find that broad fine-tuning on human choices makes task processes more human-like relative to off-the-shelf frontier agents, and task-specific process-level fine-tuning further improves human-like behavioral mimicry, though this advantage largely disappears under cross-task transfer when process targets do not naturally generalize across tasks. These results highlight process specification as a central bottleneck in achieving human-like cognitive processes in machines.
\end{abstract}

\section{Introduction}
\label{sec:intro}

Reliable human-machine discrimination is becoming increasingly important as large language models and autonomous agents are deployed in online settings, creating new challenges for security, fraud prevention, and platform integrity \citep{weidinger2022taxonomy,yao2024survey,park2024ai}. Existing approaches to distinguishing humans from machines are largely task performance-based, evaluating whether a system can produce behavior or responses indistinguishable from those of a human. This paradigm underlies systems such as CAPTCHA-style challenge-response tests \citep{von2003captcha} and automated content detectors, and reflects the enduring influence of the Turing Test, which answered ``Can machines think?'' by testing whether machines could produce human-like output \citep{turing1950computing}.

Modern AI systems increasingly satisfy this output-based bar. GPT-4.5 is judged human 73\% of the time in controlled Turing Tests \citep{jones2025large}, frontier vision models can solve reCAPTCHAv2 challenges \citep{plesner2024breaking}, and specialized AI systems now match or exceed human experts in domains once considered canonical tests of intelligence, including chess and Go \citep{silver2017mastering,campbell2002deep}. As AI systems approach human-level performance on many output-based evaluations, output is an increasingly weak criterion for distinguishing humans and machines.

When humans and AI systems produce similar task outputs, the latent processes generating those outputs may differ substantially. Artificial systems can arrive at the same answers through strategies that diverge from those used by humans, including reliance on spurious shortcuts, non-robust features, or memorized mappings \citep{geirhos2020shortcut,lapuschkin2019unmasking,block1981psychologism,regan2014human}. This motivates examining whether process-level behavioral features provide discriminative information beyond task performance and outputs alone. Because such signatures reflect latent constraints and strategies governing how decisions are generated rather than merely whether a task is solved successfully, they may remain informative even when human and machine outputs converge. We call this framework a \textit{Process Turing Test}.

To leverage process traces for distinguishing humans and agents, we lean on insights and tools from cognitive psychology, a field that has long used structured behavioral tasks to infer underlying cognitive mechanisms from measurable regularities in behavior. Examples include working memory limitations \citep{cowan2001magical,collins2012much}, systematic trial-to-trial adaptation following errors \citep{rabbitt1966errors}, and individual differences in sensitivity to wins and losses \citep{nowak1993strategy}. These patterns provide candidate \emph{process-level features} that can distinguish humans from agents even when task performance (i.e., accuracy) is matched. Furthermore, cognitive psychology provides a wealth of tasks validated over decades of research, with established task-dependent cognitive processes we can leverage to explore differences between humans and agents.

However, identifying informative process-level signatures is only part of the challenge: for such signals to serve as robust discriminators, they must remain difficult for agents to reproduce even under targeted adversarial mimicry. Modern AI models are highly adaptable, and can be optimized through fine-tuning or reinforcement learning towards target behavioral objectives, including alignment with human preferences and response styles \citep{ouyang2022training,bai2022constitutional,rafailov2023direct,binz2025foundation}. This raises a deeper question: even if process-level features distinguish humans from agents, how robust does this distinction remain after targeted adaptation?

We structure our findings in two stages:

\textbf{First, how much additional discriminative value does process provide beyond output alone?}
To evaluate this, we introduce a battery of 30 cognitive tasks drawn from cognitive psychology with extensively operationalized process-level behavioral signatures. These tasks were selected to leverage well-characterized differences in \emph{how} humans solve problems—not merely whether they solve them—thereby providing a rich testbed for comparing human and agent behavior. Across the battery, we find that process-level behavioral features provide substantially greater discriminative signal than performance metrics alone (e.g., accuracy, earned points), revealing systematic human-agent differences even when task performance is similar (random forest classifier AUC = 0.88 across performance matched tasks).

\textbf{Second, what limits an agent's ability to reproduce human-like process?}
We compare frontier agents (Claude Sonnet 4.5, GPT-5, Gemini 2.5 Pro), Centaur \citep{binz2025foundation}, a 70B model fine-tuned on 10.7M human decisions across 160+ tasks as an approximate upper bound on human-like behavior among current models. We then progressively fine-tune Qwen2.5-Instruct (1.5B, 7B, 14B, 32B) and Llama3.3-Instruct (70B) using task-specific imitation of individual human actions (\textbf{action-level supervised fine-tuning} (\textbf{A-SFT})) and direct task-level optimization of behavioral process features (\textbf{process-level fine-tuning} (\textbf{P-SFT})). We find that A-SFT substantially improves human-like process relative to off-the-shelf frontier agents, reaching a mean process distance of $\lvert d \rvert = 0.40$, comparable to Centaur ($\lvert d \rvert = 0.44$). When task-specific process representations are known and aligned to the evaluation task, P-SFT further reduces this distance to $\lvert d \rvert = 0.22$. However, this advantage largely disappears under cross-task transfer, where P-SFT and A-SFT reach mean distances of $\lvert d \rvert = 0.66$ and $\lvert d \rvert = 0.62$, respectively. %



Together, our results show that process equivalence is independent of output equivalence, and we identify generalization beyond explicit process representations as a central challenge for scalable human-process alignment.

\section{Output Equivalence vs. Process Equivalence}

We constructed a battery of 30 tasks largely drawn from cognitive psychology. The battery includes tasks such as mental rotation \citep{shepard1971mental} and the Stroop task \citep{stroop1935studies} -- classical psychology tasks with extensively operationalized and validated process measures (Figure \ref{fig:current_results}).

We begin by describing the shared human-agent evaluation setup, then use the classical CAPTCHA as an intuitive real-world example motivating the broader benchmark, followed by the full-battery evaluation of agents and humans.

\subsection{Participants and AI Agents}

We administered the 30-task battery to 100 human participants recruited via Prolific (all participants provided IRB-approved consent; 97 were retained after excluding runs affected by platform issues) and collected 50 full-battery runs from each of three frontier vision-language agents (GPT-5, Gemini 2.5 Pro, and Claude Sonnet 4.5; 150 total agent runs). Agents interacted with the same browser-based interfaces shown to human participants. On each trial, the agent received a screenshot and returned a structured JSON action (e.g., target coordinates, tile index, or categorical choice), which a Playwright-based execution layer translated into the corresponding DOM event (click, keypress, or input). Example task instructions for both agents and participants are provided in Appendix~\ref{app:task_prompts}.

\subsection{CAPTCHA as a Motivating Real-World Example}
\label{sec:captcha_example}

\begin{figure}[ht]
    \centering
    \includegraphics[width=1\linewidth]{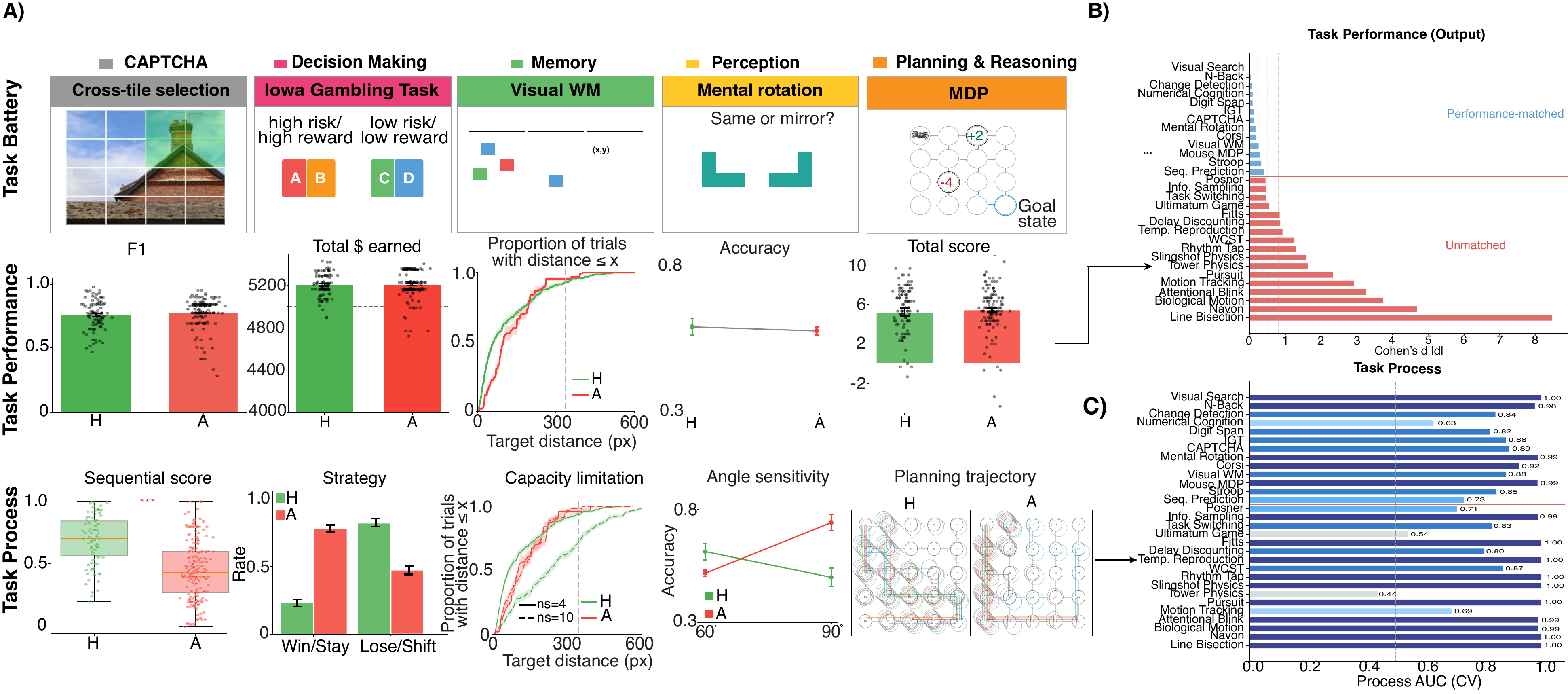}
\caption{
\textbf{Process-level behavioral features provide stronger human-agent discriminative signal than task performance alone.}
\textbf{A)} The task battery consists of 30 tasks spanning decision making, memory, perception, and planning/reasoning, each paired with task-specific process features designed to capture how behavior unfolds over time. Representative examples illustrate matched task performance despite divergent process-level signatures across multiple domains. Error bars represent standard errors, individual dots represent individual human or agent participants.
\textbf{B)} Per-task output distance between humans and agents measured by absolute Cohen's $d$ on each task's primary performance metric (e.g., accuracy, score, reward). Tasks above the red threshold are considered performance-matched. Many tasks show near-human agent performance.
\textbf{C)} Per-task cross-validated AUC of a Random Forest classifier trained to distinguish humans from agents using process-level features. Despite good performance matching, process-level features remain highly discriminative across most tasks, indicating that output equivalence does not imply process equivalence.
}
\label{fig:current_results}
\end{figure}

We begin with CAPTCHA, a real-world human-machine discrimination task. Modern AI systems can solve many CAPTCHA variants at near-human accuracy \citep{plesner2024breaking}. However, matching CAPTCHA performance does not imply matching the behavioral process by which the challenge is solved. To illustrate this, we compare humans and agents on two CAPTCHA variants: a \emph{Classic} format in which participants select target images from a grid, and a \emph{Cross-Tile} format in which participants identify all tiles containing a target object within a single image. Humans and agents achieve statistically indistinguishable performance on these tasks ($t(98)=-0.29$, $p=0.77$, Figure~\ref{fig:current_results}A), indicating matched performance. However, their interaction patterns differ substantially: process-level features such as sequential click patterns and column/row switches reliably distinguish human from agent CAPTCHA behavior (Figure~\ref{fig:current_results}A). Thus, output-matched agents need not be process-matched.

While CAPTCHA provides an intuitive real-world example of performance/process dissociation, it captures only a narrow and domain-specific slice of human behavior. We therefore extend this analysis across the cognitive psychology corpus to the remaining tasks in our battery, which span a broad range of cognitive domains and permit systematic measurement of process-level human-agent differences.


\subsection{The cognitive task battery}

\subsubsection{Task selection}

The remaining tasks in the battery are classical cognitive psychology paradigms spanning working memory, decision-making, perception, planning, and reasoning (Table \ref{tab:CogCAPTCHA30_feature_catalog}). We selected these tasks based on three criteria. First, the tasks span a broad range of cognitive domains with minimal overlap in the processes they measure (Fig. \ref{fig:feature-correlation-matrix}). Second, all tasks are well-validated paradigms with established literature characterizing their underlying cognitive processes and associated process measures. Canonical sources for each task are provided in Table \ref{tab:CogCAPTCHA30_feature_catalog} (e.g., \citep{corsi1972human,fitts1954information,shepard1971mental}). Third, the tasks were configured to require at most 10 trials and less than one minute to complete, keeping their duration comparable to that of a CAPTCHA. The complete set of tasks can be viewed at \url{https://d1ez37832ply4h.cloudfront.net/}.

\subsubsection{Process feature selection}

Process features were drawn per task from the established process measures of each paradigm's literature. Decision-making tasks capture strategy signatures such as win-stay/lose-shift behavior, choice stickiness, delay discounting rates, and early vs. late exploration (e.g., Iowa Gambling Task, delay discounting); working-memory tasks capture capacity and serial-position signatures such as set-size sensitivity, primacy/recency effects, and transposition errors (e.g., digit span, Corsi); perceptual tasks capture bias and precision signatures such as side bias, distance-dependent error, and congruency effects (e.g., line bisection, visual search, Stroop); planning/reasoning tasks capture information-gathering and rule-updating signatures such as pre-decision exploration and perseveration (e.g., Mouselab-MDP, Wisconsin Card Sorting Task).

\subsection{Process Features Reveal Human-Agent Differences}

All human and agent runs were transformed into a common feature vector comprising a single primary performance output metric (e.g. accuracy, total score), and multiple validated task-specific process features. These process features were collected across the 30 tasks and extracted using task-specific featurizers that we applied uniformly to humans and agents (Appendix Table~\ref{tab:CogCAPTCHA30_feature_catalog}). Examples of processes and features are shown in Figure \ref{fig:current_results}A.

On output metrics, agents achieved statistical parity with humans on 13 of 30 tasks (Mann--Whitney U, $p \geq 0.05$; median Cohen's $\lvert d \rvert = 0.08$ on matched tasks vs.\ 1.29 on unmatched tasks; Figure \ref{fig:current_results}B), defining a subset of tasks where output-based discrimination would fail.

Notably, many of the tasks on which agents underperformed humans involved continuous visual stimuli (e.g., biological motion where continuous human figure movement implies direction, and other video-based paradigms). This likely reflects a limitation of the evaluated vision-language agents, which typically operate over discrete screen snapshots instead of continuous visual streams, rather than a fundamental inability to perform such tasks. As multimodal systems gain stronger support for continuous visual input, performance gaps on these tasks will likely diminish. Therefore, we consider inclusion of these tasks to be appropriate.

While agents achieved output parity with humans on 13 tasks, process-level features reliably distinguished humans from agents regardless of whether their final outputs were matched. A per-task Random Forest classifier trained to distinguish humans from agents using process features (5-fold stratified cross-validation per task; 200 trees; max depth 5; class-balanced) achieved a mean AUC of 0.88 across output-matched tasks (Figure~\ref{fig:current_results}C), comparable to the mean AUC of 0.87 on output-unmatched tasks ($n = 17$). Repeating the same Random Forest analysis using output metrics alone yielded substantially lower discriminative performance (mean AUC = 0.55 on output-matched tasks, mean AUC = 0.80 on output-unmatched tasks, and mean AUC = 0.77 across all tasks). Together, these results indicate that process-level behavioral features provide substantially stronger discriminative signal than output metrics alone, and remain informative even when human and agent task performance converges.

\section{Closing the Process Gap Through Action Imitation and Supervised Fine-tuning}

The preceding results establish that process-level behavioral features reliably distinguish humans from current AI agents even when task performance is matched. The natural next question is what limits an agent's ability to reproduce human-like process. In particular, is the observed process gap primarily due to insufficient exposure to human behavioral data, and/or does closing it require more direct and task-specific forms of supervision? We first compare frontier models to Centaur, a foundation model of human cognition \citep{binz2025foundation}, and then fine-tune models (Qwen2.5-Instruct (1.5B, 7B, 14B, 32B) and Llama3.3-Instruct 70B, selected to match Centaur's model family and scale) using progressively more targeted task and process supervision.


\subsection{Centaur: A Foundation Model of Human Cognition \texorpdfstring{\citep{binz2025foundation}}{}}

While off-the-shelf language models are not explicitly optimized to reproduce human behavior in structured cognitive tasks, Centaur \citep{binz2025foundation} is a Llama3.1-70B foundation model trained via supervised fine-tuning to predict human decisions using over 10 million choices from more than 160 cognitive experiments. Centaur often outperforms domain-specific cognitive models designed to capture human behavior, making it a strong benchmark for evaluating the extent to which broad human-choice imitation can recover human-like cognitive processes.

Centaur operates exclusively on text-based prompts so we restrict our subsequent analysis to three tasks with a structured trial format that can be readily represented as text: the Iowa Gambling Task (IGT), Wisconsin Card Sorting Task (WCST), and Information Sampling Task (task descriptions and respective features are provided in Fig. \ref{fig:distance}A). This set of three tasks allows us to directly evaluate Centaur with frontier AI agents and our fine-tuned models.

\subsection{Evaluation Metrics}

We evaluate agents on these tasks using two complementary metrics:

\textbf{Distributional distance.} For each model and task, we computed Cohen's $\lvert d \rvert $ between model and human feature distributions for each process feature, and report mean $\lvert d \rvert$ across features together with energy distance between joint distributions \citep{szekely2013energy}; energy results are reported in Appendix Table \ref{tab:energy}, and lower values indicate closer alignment to human process patterns.

\textbf{Classifier fool rate.} We trained a separate Random Forest classifier (200 trees; max depth 5; class-balanced) using only features from the three evaluation tasks. The classifier was trained to distinguish 97 humans from eight models: three frontier AI agents and the base models at each of the five model sizes. Held-out runs from these models were reserved for evaluation, while Centaur and all fine-tuned models were excluded from classifier training. We report each model's fool rate: the fraction of held-out runs assigned $P(\mathrm{human}) \geq 0.5$, along with the distribution of classifier-assigned $P(\mathrm{human})$ values.

Centaur matched human process distributions more closely than frontier agents across the three evaluated tasks (Fig. \ref{fig:distance}B), achieving substantially lower mean distributional distance (Centaur: $\lvert d \rvert = 0.44$; Gemini: $\lvert d \rvert = 1.07$; GPT: $\lvert d \rvert = 1.36$; Claude: $\lvert d \rvert = 1.65$). Likewise, Centaur achieved substantially higher classifier fool rates than all frontier agents (Centaur fool rate: 44\%, remaining models $\leq$ 20\%; Centaur $\mathrm{p(Human)}$: 0.45, remaining models $\mathrm{p(Human)}$ $\leq$ 0.20). Full, task-specific results are reported in Table \ref{tab:cross_task}. The narrowing of the process gap from frontier agents to Centaur suggests that large-scale training on human choice data can shift model behavior toward human-like cognitive processes.


\subsection{Supervised Fine-tuning}
\label{sec:adversarial}

Results in Fig. \ref{fig:current_results} demonstrate that process-level behavioral features can distinguish humans from agents even when task outputs are matched, and our comparison to Centaur shows that fine-tuning on large-scale human action data can partially narrow this gap. However, these results do not imply that action imitation represents the upper bound of human-like process mimicry. Human-machine discrimination is inherently an \emph{adversarial} problem, and thus any behavioral feature used to distinguish humans from agents may itself become a target for optimization. 

This motivates a stronger test: can an agent close the process gap when given increasingly direct access to human cognitive process supervision? We test this through two fine-tuning interventions, evaluating each stage in the same way we evaluated Centaur and the frontier agents: (1) whether the resulting process-feature distributions move closer to the human distribution, and (2) whether a process-based classifier can still distinguish the modified agent from humans.

We tested two forms of fine-tuning:

    \textbf{Task-specific action-level fine-tuning (A-SFT).} We fine-tuned open-source language models spanning different sizes and model families (Qwen2.5-Instruct - sizes 1.5B, 7B, 14B, 32B and Llama3.3-Instruct 70B model, to match Centaur's architecture) via LoRA \citep{hu2021lora} to predict the human action at each individual decision point. This aligns the model to per-decision human action targets but imposes no explicit objective over run-level behavioral patterns: small per-trial deviations can accumulate across a session, yielding aggregate process features that diverge from the human population. This intervention tests whether \textit{task-specific} action imitation (rather than the large-scale imitation provided by Centaur) is sufficient to reproduce human-like process.
            
    \textbf{Process-level fine-tuning (P-SFT).} The same set of models was fine-tuned with an additional loss term that directly aligned the run-level process-feature distribution with that of the human population. 
    


At each training step, in addition to the standard cross-entropy loss on individual human actions, we ran the model forward through a complete simulated run of the task and computed its expected process feature vector under its own current policy. The full training objective was:
\begin{equation}
\mathcal{L} = \mathcal{L}_{\text{CE}} + \lambda_{\text{diff}} \sum_k w_k \left( \frac{f_k^{\text{model}} - \mu_k^{\text{human}}}{\sigma_k^{\text{human}}} \right)^2,
\label{eq:psft_loss}
\end{equation}
where $\mathcal{L}_{\text{CE}}$ is the standard SFT cross-entropy on human action choices, $f_k^{\text{model}}$ is the differentiable estimate of run process feature $k$ under the current policy, $(\mu_k^{\text{human}}, \sigma_k^{\text{human}})$ are the human-population mean and standard deviation of process feature $k$, and $w_k$ allows up-weighting features that converge more slowly. $\lambda$ controlled the relative weight of the feature-matching term against the action-level cross-entropy (with $\lambda$ = 1 meaning equal contribution; hyperparameters are given in Appendix~\ref{app:psft_details}).

\paragraph{P-SFT illustrative example: Iowa Gambling Task.}
To make the P-SFT mechanism concrete, we illustrate it using the Iowa Gambling Task (IGT), a task in which participants choose among four card decks (A--D) over 10 trials (the same procedure was also applied to the Wisconsin Card Sorting Task (WCST) and Information Sampling Task in our broader fine-tuning experiments, Appendix \ref{app:psft_details}). The targeted process features are learning slope, stickiness, deck entropy, win-stay rate, lose-shift rate, and good-deck rate. On each trial, the model produces a probability distribution over the four decks, e.g. $p(A)=0.15$, $p(B)=0.10$, $p(C)=0.30$, $p(D)=0.45$. These probabilities feed directly into the feature estimates without waiting to see which deck the model actually picks.

For instance, if the model selected deck C on the previous trial, its stickiness estimate for the current trial is simply $p(C)=0.30$ — the probability of repeating the previous choice. Its good-deck-rate estimate is $p(C)+p(D)=0.75$, since decks C and D are the less risky decks. At the end of the run, per-trial estimates are averaged into run-level features (e.g., mean stickiness across all trials). To construct the prompt for the next trial, one deck is sampled from this distribution and the corresponding outcome is appended to the context (e.g., ``You chose deck C and won 50.\ Balance: 5050.''). This sampled action shapes the subsequent prompt but is not differentiated through: gradients flow only through the probability-based feature estimates described above. The mismatch between the resulting run-level features and human means is then penalized via Equation~\ref{eq:psft_loss}. 

We also tested whether any similarity yielded by process-level supervision generalizes beyond the specific behavioral representations used during optimization. We withheld additional process features from the optimization objective and reserved them for evaluation only. These left-out features span distinct behavioral dimensions not directly optimized during training, allowing us to assess whether improvements transfer beyond the supervised process representation (Table \ref{tab:feature_combinations} shows consistent results across different optimization-holdout feature combinations). Additionally, we performed a cross-task evaluation in which models were assessed on process features from tasks outside the supervised fine-tuning task. We report left-out-feature and cross-task results separately in the following section (See Fig. \ref{fig:classifier-fool} for details).


\subsection{Results}

\begin{figure}[t!]
    \centering
    \includegraphics[width=0.9\linewidth]{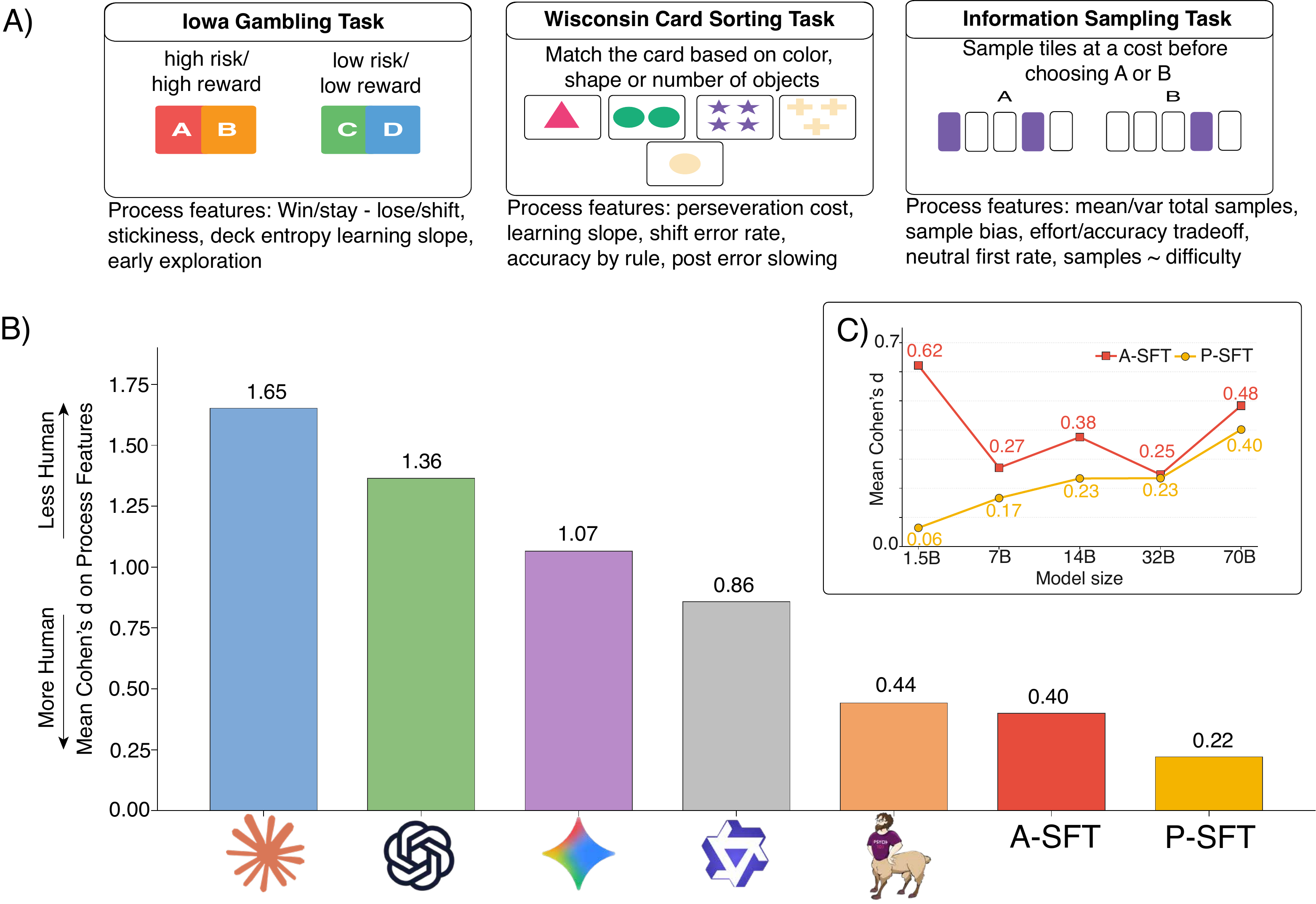 }
\caption{\textbf{Process-supervised fine-tuning produces the most human-like cognitive process signatures}. A) The three decision-making tasks and their associated process features. In the Iowa Gambling Task, participants choose repeatedly among four decks that differ in risk/reward structure; in the Wisconsin Card Sorting Task, participants infer a hidden matching rule (color, shape, or number) that changes without warning; in the Information Sampling Task, participants reveal tiles at a cost before judging which of two options has higher values. Process features capture how decisions unfold (e.g., win-stay/lose-shift, perseveration cost, sampling effort) rather than final outcomes. B) Distance to human behavior, measured as the mean absolute Cohen's d between model and human distributions over process features (averaged within each task, then across the three tasks; lower = more human-like). C) Mean process-feature distance for A-SFT and P-SFT across model sizes.}
\label{fig:distance}
\end{figure}

\textbf{Distributional distance.}
As stated above, off-the-shelf LLMs (Claude, GPT-5, Gemini, and base Qwen 1.5B) show large deviations from human process signatures ($\lvert d \rvert = 0.86$--$\lvert d \rvert = 1.65$). Centaur, a 70B model fine-tuned on human choice data, substantially reduces this distance ($\lvert d \rvert = 0.44$). Action-supervised fine-tuning (A-SFT; trained to imitate task-specific human action sequences) and process-supervised fine-tuning (P-SFT; trained to match human process-feature distributions) further reduce the gap across five model sizes. A-SFT achieves a mean distance of $\lvert d \rvert = 0.40$, comparable to Centaur, whereas P-SFT yields the closest match to human process signatures ($\lvert d \rvert = 0.22$; Fig. \ref{fig:distance}B). Across individual model sizes, P-SFT outperforms A-SFT at every scale, with the largest difference at 1.5B parameters ($\lvert d \rvert = 0.06$ vs. $\lvert d \rvert = 0.62$).

Together, these results suggest that explicit process-level supervision can substantially improve human-like behavioral imitation when the relevant process representation is known and aligned with the target task. However, as we examine below, these gains diminish when models are evaluated on tasks with non-overlapping process-feature spaces. The full set of results, including task-specific results, is provided in Table \ref{tab:cross_task} and Figure \ref{fig:distance-all-models}.

\textbf{Classifier fool rates}

\begin{figure}[t!]
    \centering
    \includegraphics[width=1\linewidth]{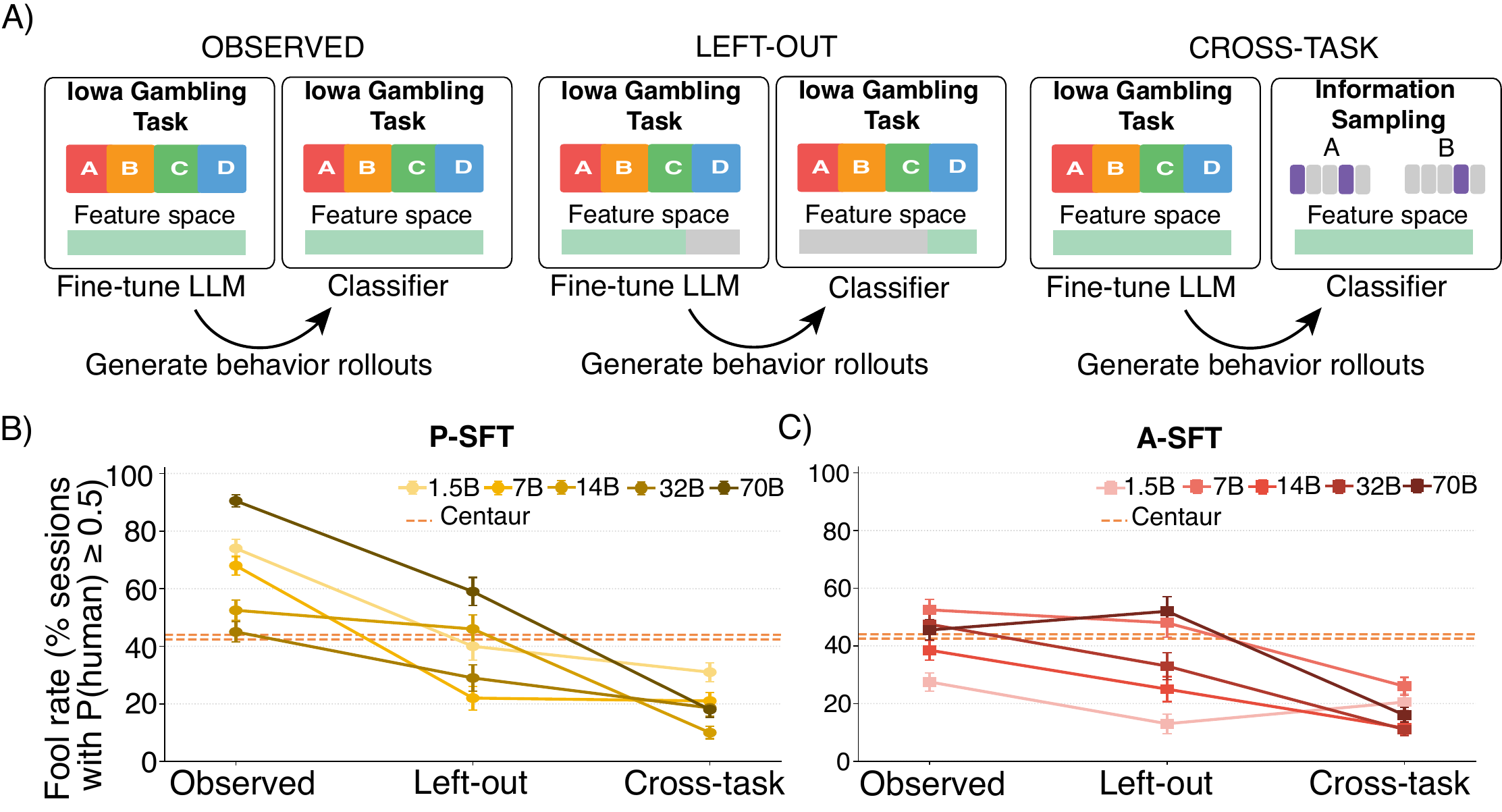 }
\caption{\textbf{Human-likeness from process-supervised fine-tuning is specific to the observed (supervised) features and task}. A) Three evaluation regimes for fine-tuned models. Observed: rollouts are generated on the training task and evaluated on the same process features used during fine-tuning. Left-out: rollouts on the training task are evaluated on left-out process features never used in fine-tuning. Cross-task: the fine-tuned model generates rollouts on a different task (e.g., fine-tuned on the Iowa Gambling Task, evaluated on Information Sampling), scored on that task's features. The percentage of sessions a Random Forest classifier scores as human (P(human) $\geq$ 0.5) for P-SFT (B) and A-SFT (C) at each model size, across the three regimes. Error bars show binomial SEM; the dashed line marks Centaur (44\% on matched observed features, 43\% on matched left-out features). P-SFT achieves high fool rates on observed features (up to 90.5\% at 70B), but performance degrades on left-out features and collapses to 10–31\% cross-task at every size, indicating that process alignment partially transfers to unmeasured features within the same task but does not generalize reliably to novel tasks.}
   
\label{fig:classifier-fool}
\end{figure}

Results from the held-out Random Forest classifier mirrored the distributional-distance analysis. Off-the-shelf frontier agents were reliably detected, with fool rates of $\leq 20\%$ (Table \ref{tab:fool_combined}). Centaur more than doubled the fool rate of the frontier agents, with $44\%$ of sessions classified as human. Across model sizes, A-SFT achieved a mean fool rate of $42\%$ (range: $27.5\%$--$52.5\%$), while P-SFT reached $66\%$ (range: $45\%$--$90.5\%$; Fig. \ref{fig:classifier-fool}B,C; Table \ref{tab:fool_combined}). Thus, while both Centaur and A-SFT substantially increased classifier fool rates relative to off-the-shelf agents, P-SFT produced the largest increase, further demonstrating the advantage of direct process-level supervision in producing behavior that is difficult to distinguish from human behavior.

\textbf{Generalization tests}

To test whether the advantage of P-SFT generalizes beyond the process features and tasks used during fine-tuning, we performed two complementary evaluations. First, in the \textit{left-out feature} condition, we fine-tuned each model using only a subset of process features, generated task behavior from the fine-tuned model, and evaluated it using a classifier trained on a separate, left-out set of features (Fig. \ref{fig:classifier-fool}A). These left-out features span distinct behavioral dimensions not directly optimized during training, allowing us to assess whether improvements transfer beyond the supervised process representation (Table \ref{tab:feature_combinations} shows consistent results across different optimization-holdout feature combinations). Second, in the \textit{cross-task} condition, we fine-tuned each model on one task, generated behavioral rollouts on a different task, and evaluated the resulting behavior using the classifier for that task (Fig. \ref{fig:classifier-fool}A).

In the left-out feature evaluation, P-SFT retained a modest advantage over A-SFT, although both were roughly close to Centaur (classifier fool rate on matched left-out features $43\%$) and P-SFT’s fool rate decreased substantially relative to evaluation on the features used during supervision. P-SFT achieved a mean fool rate of $39\%$ (range: $22\%$--$59\%$ across model sizes; Fig. \ref{fig:classifier-fool}B), compared with $34\%$ for A-SFT (range: $13\%$--$52\%$; Fig. \ref{fig:classifier-fool}C). In the cross-task evaluation, where models were assessed on tasks with process representations different from those used during supervision, the advantage of P-SFT largely disappeared. P-SFT achieved a mean fool rate of $20\%$ (range: $10\%$--$31\%$; Fig. \ref{fig:classifier-fool}B), compared with $17\%$ for A-SFT (range: $11\%$--$26\%$; Fig. \ref{fig:classifier-fool}C), with both falling below Centaur.

These findings suggest that process-level supervision yields its largest gains when the target process representation is directly available during training. This advantage attenuates when using left-out features within the same task, and it does not readily generalize across tasks with different process representations. Per-task results are provided in Table \ref{tab:fool_combined}.

\section{Discussion}
\label{sec:discussion}

Large language models increasingly challenge output-based notions of human-machine discrimination, motivating a shift from evaluating \textit{what} models produce, the original Turing Test criterion, to \textit{how} they produce it, what we call a \textit{Process Turing Test}. Across our cognitive task battery, process-level behavioral features provided substantially stronger discriminative signal than output metrics alone, reliably distinguishing humans from agents even when their task performance was comparable.

We next asked whether this process gap persists as models are increasingly optimized to imitate human behavior. Centaur, a foundation model trained on large-scale human choice data, substantially narrowed the gap relative to frontier agents. Task-specific action imitation (A-SFT) across Qwen2.5 and Llama3.3 produced similar gains, showing that direct imitation of human actions can shift model behavior toward human process signatures. Explicit process-level supervision (P-SFT) narrowed the gap further, producing the closest match to human behavior when the relevant task-specific process representation was directly accessible/observed. However, this advantage diminished for left-out features and largely disappeared under cross-task evaluation, indicating that matching a known process representation does not necessarily produce more general human-like behavior.

Together, these results suggest that process provides more discriminative power than outputs across current AI models, and that broader human-process alignment will require representations that capture transferable structure across tasks rather than task-specific feature matching alone. Such representations could take the form of latent behavioral embeddings, hierarchical behavioral abstractions, or multi-task objectives that capture shared structure across different cognitive domains.

\paragraph{Limitations.}

Several limitations remain. First, our human data were drawn from a relatively small participant sample from a single pool. We therefore operationalize ``human-like'' process relative to the empirical behavioral distribution of this sampled population under the chosen feature representation, rather than as a claim about any canonical form of human cognition. Second, our fine-tuning evaluation is restricted to a limited subset of our original task battery, as we aimed to benchmark the methods against Centaur which only accepts text data in a formatted trial structure. Accordingly, it remains unclear whether similar process-alignment dynamics will hold for richer interactive environments. Lastly, process-based discrimination need not depend on a fixed battery of behavioral probes. Automated task generation and iterative task redesign could continuously introduce novel behavioral demands and target dimensions along which human and machine processes remain distinct, turning process-based evaluation into a dynamic red-teaming framework in which increasingly human-like models are evaluated against new process-level challenges rather than a static set of known behavioral signatures.



\bibliography{iclr2027_conference}

@article{binz2025foundation,
  title={A foundation model to predict and capture human cognition},
  author={Binz, Marcel and Akata, Elif and Bethge, Matthias and Br{\"a}ndle, Franziska and Callaway, Fred and Coda-Forno, Julian and Dayan, Peter and Demircan, Can and Eckstein, Maria K and {\'E}ltet{\H{o}}, No{\'e}mi and others},
  journal={Nature},
  volume={644},
  number={8078},
  pages={1002--1009},
  year={2025},
  publisher={Nature Publishing Group UK London}
}

@article{collins2012much,
  title={{How much of reinforcement learning is working memory, not reinforcement learning? A behavioral, computational, and neurogenetic analysis}},
  author={Collins, Anne GE and Frank, Michael J},
  journal={European Journal of Neuroscience},
  volume={35},
  number={7},
  pages={1024--1035},
  year={2012},
  publisher={Wiley Online Library}
}

@article{cowan2001magical,
  title={{The magical number 4 in short-term memory: A reconsideration of mental storage capacity}},
  author={Cowan, Nelson},
  journal={Behavioral and Brain Sciences},
  volume={24},
  number={1},
  pages={87--114},
  year={2001},
  publisher={Cambridge University Press}
}

@article{rabbitt1966errors,
  title={Errors and error correction in choice-response tasks},
  author={Rabbitt, Patrick M. A.},
  journal={Journal of Experimental Psychology},
  volume={71},
  number={2},
  pages={264--272},
  year={1966}
}

@article{turing1950computing,
  title={{Computing Machinery and Intelligence}},
  author={Turing, Alan M.},
  journal={Mind},
  volume={59},
  number={236},
  pages={433--460},
  year={1950}
}

@article{block1981psychologism,
  title={{Psychologism and Behaviorism}},
  author={Block, Ned},
  journal={The Philosophical Review},
  volume={90},
  number={1},
  pages={5--43},
  year={1981},
  publisher={JSTOR}
}

@article{hu2021lora,
  title={{LoRA: Low-Rank Adaptation of Large Language Models}},
  author={Hu, Edward J. and Shen, Yelong and Wallis, Phillip and Allen-Zhu, Zeyuan and Li, Yuanzhi and Wang, Lu and Wang, Weizhu and Chen, Weizhu},
  journal={arXiv preprint arXiv:2106.09685},
  year={2021}
}

@article{szekely2013energy,
  title={{Energy statistics: A class of statistics based on distances}},
  author={Sz{\'e}kely, G{\'a}bor J and Rizzo, Maria L},
  journal={Journal of Statistical Planning and Inference},
  volume={143},
  number={8},
  pages={1249--1272},
  year={2013},
  publisher={Elsevier}
}

@article{qwen2.5,
  title={{Qwen2.5 Technical Report}},
  author={Qwen Team},
  journal={arXiv preprint arXiv:2412.15115},
  year={2024}
}

@article{silver2017mastering,
  title={{Mastering the game of Go without human knowledge}},
  author={Silver, David and Schrittwieser, Julian and Simonyan, Karen and Antonoglou, Ioannis and Huang, Aja and Guez, Arthur and Hubert, Thomas and Baker, Lucas and Lai, Matthew and Bolton, Adrian and others},
  journal={Nature},
  volume={550},
  number={7676},
  pages={354--359},
  year={2017}
}

@inproceedings{plesner2024breaking,
  title={{Breaking reCAPTCHAv2}},
  author={Plesner, Andreas and Vontobel, Tobias and Wattenhofer, Roger},
  booktitle={2024 IEEE 48th Annual Computers, Software, and Applications Conference (COMPSAC)},
  pages={1047--1056},
  year={2024},
  organization={IEEE}
}

@article{jones2025large,
  title={{Large Language Models Pass the Turing Test}},
  author={Jones, Cameron R and Bergen, Benjamin K},
  journal={arXiv preprint arXiv:2503.23674},
  year={2025}
}

@article{nowak1993strategy,
  title={{A strategy of win-stay, lose-shift that outperforms tit-for-tat in the Prisoner's Dilemma game}},
  author={Nowak, Martin and Sigmund, Karl},
  journal={Nature},
  volume={364},
  number={6432},
  pages={56--58},
  year={1993},
  publisher={Nature Publishing Group UK London}
}

@article{campbell2002deep,
  title={{Deep Blue}},
  author={Campbell, Murray and Hoane Jr, A Joseph and Hsu, Feng-hsiung},
  journal={Artificial Intelligence},
  volume={134},
  number={1-2},
  pages={57--83},
  year={2002},
  publisher={Elsevier}
}

@article{geirhos2020shortcut,
  title={Shortcut learning in deep neural networks},
  author={Geirhos, Robert and Jacobsen, J{\"o}rn-Henrik and Michaelis, Claudio and Zemel, Richard and Brendel, Wieland and Bethge, Matthias and Wichmann, Felix A},
  journal={Nature Machine Intelligence},
  volume={2},
  number={11},
  pages={665--673},
  year={2020},
  publisher={Nature Publishing Group UK London}
}

@article{lapuschkin2019unmasking,
  title={{Unmasking Clever Hans predictors and assessing what machines really learn}},
  author={Lapuschkin, Sebastian and W{\"a}ldchen, Stephan and Binder, Alexander and Montavon, Gr{\'e}goire and Samek, Wojciech and M{\"u}ller, Klaus-Robert},
  journal={Nature Communications},
  volume={10},
  number={1},
  pages={1096},
  year={2019},
  publisher={Nature Publishing Group UK London}
}

@article{ouyang2022training,
  title={Training language models to follow instructions with human feedback},
  author={Ouyang, Long and Wu, Jeffrey and Jiang, Xu and Almeida, Diogo and Wainwright, Carroll and Mishkin, Pamela and Zhang, Chong and Agarwal, Sandhini and Slama, Katarina and Ray, Alex and others},
  journal={Advances in Neural Information Processing Systems},
  volume={35},
  pages={27730--27744},
  year={2022}
}

@article{bai2022constitutional,
  title={{Constitutional AI: Harmlessness from AI Feedback}},
  author={Bai, Yuntao and Kadavath, Saurav and Kundu, Sandipan and Askell, Amanda and Kernion, Jackson and Jones, Andy and Chen, Anna and Goldie, Anna and Mirhoseini, Azalia and McKinnon, Cameron and others},
  journal={arXiv preprint arXiv:2212.08073},
  year={2022}
}

@article{rafailov2023direct,
  title={{Direct Preference Optimization: Your Language Model is Secretly a Reward Model}},
  author={Rafailov, Rafael and Sharma, Archit and Mitchell, Eric and Manning, Christopher D and Ermon, Stefano and Finn, Chelsea},
  journal={Advances in Neural Information Processing Systems},
  volume={36},
  pages={53728--53741},
  year={2023}
}

@article{yao2024survey,
  title={{A survey on large language model (LLM) security and privacy: The Good, The Bad, and The Ugly}},
  author={Yao, Yifan and Duan, Jinhao and Xu, Kaidi and Cai, Yuanfang and Sun, Zhibo and Zhang, Yue},
  journal={High-Confidence Computing},
  volume={4},
  number={2},
  pages={100211},
  year={2024},
  publisher={Elsevier}
}

@inproceedings{weidinger2022taxonomy,
  title={{Taxonomy of Risks posed by Language Models}},
  author={Weidinger, Laura and Uesato, Jonathan and Rauh, Maribeth and Griffin, Conor and Huang, Po-Sen and Mellor, John and Glaese, Amelia and Cheng, Myra and Balle, Borja and Kasirzadeh, Atoosa and others},
  booktitle={Proceedings of the 2022 ACM Conference on Fairness, Accountability, and Transparency},
  pages={214--229},
  year={2022}
}

@article{park2024ai,
  title={{AI deception: A survey of examples, risks, and potential solutions}},
  author={Park, Peter S and Goldstein, Simon and O’Gara, Aidan and Chen, Michael and Hendrycks, Dan},
  journal={Patterns},
  volume={5},
  number={5},
  year={2024},
  publisher={Elsevier}
}

@inproceedings{von2003captcha,
  title={{CAPTCHA: Using hard AI problems for security}},
  author={Von Ahn, Luis and Blum, Manuel and Hopper, Nicholas J and Langford, John},
  booktitle={International Conference on the Theory and Applications of Cryptographic Techniques},
  pages={294--311},
  year={2003},
  organization={Springer}
}

@inproceedings{bechara2001neurobiology,
  title={Neurobiology of decision-making: risk and reward.},
  author={Bechara, Antoine},
  booktitle={Seminars in Clinical Neuropsychiatry},
  volume={6},
  number={3},
  pages={205--216},
  year={2001}
}

@article{shapiro1997attentional,
  title={The attentional blink},
  author={Shapiro, Kimron L and Raymond, Jane E and Arnell, Karen M},
  journal={Trends in Cognitive Sciences},
  volume={1},
  number={8},
  pages={291--296},
  year={1997},
  publisher={Elsevier}
}

@article{troje2002decomposing,
  title={{Decomposing biological motion: A framework for analysis and synthesis of human gait patterns}},
  author={Troje, Nikolaus F},
  journal={Journal of Vision},
  volume={2},
  number={5},
  pages={2--2},
  year={2002},
  publisher={The Association for Research in Vision and Ophthalmology}
}

@article{luck1997capacity,
  title={The capacity of visual working memory for features and conjunctions},
  author={Luck, Steven J and Vogel, Edward K},
  journal={Nature},
  volume={390},
  number={6657},
  pages={279--281},
  year={1997},
  publisher={Nature Publishing Group UK London}
}

@article{corsi1972human,
  title={{Human Memory and the Medial Temporal Region of the Brain.}},
  author={Corsi, Philip Michael},
  year={1972},
  publisher={McGill University}
}

@incollection{mazur2013adjusting,
  title={{An Adjusting Procedure for Studying Delayed Reinforcement}},
  author={Mazur, James E},
  booktitle={The Effect of Delay and of Intervening Events on Reinforcement Value},
  pages={55--73},
  year={2013},
  publisher={Psychology Press}
}

@book{wechsler1955wechsler,
  author    = {Wechsler, David},
  title     = {Manual for the Wechsler Adult Intelligence Scale},
  publisher = {The Psychological Corporation},
  address   = {New York},
  year      = {1955}
}

@article{fitts1954information,
  title={The information capacity of the human motor system in controlling the amplitude of movement.},
  author={Fitts, Paul M},
  journal={Journal of Experimental Psychology},
  volume={47},
  number={6},
  pages={381},
  year={1954},
  publisher={American Psychological Association}
}

@inproceedings{callaway2017mouselab,
    title={{Mouselab-MDP: A new paradigm for tracing how people plan}},
    author={Callaway, Frederick and Lieder, Falk and Krueger, Paul M and Griffiths, Thomas L},
  booktitle={The 3rd Multidisciplinary Conference on Reinforcement Learning and Decision Making, Ann Arbor, MI},
  year={2017}
}

@article{jewell2000pseudoneglect,
  title={Pseudoneglect: a review and meta-analysis of performance factors in line bisection tasks},
  author={Jewell, George and McCourt, Mark E},
  journal={Neuropsychologia},
  volume={38},
  number={1},
  pages={93--110},
  year={2000},
  publisher={Elsevier}
}

@article{shepard1971mental,
  title={{Mental Rotation of Three-Dimensional Objects}},
  author={Shepard, Roger N and Metzler, Jacqueline},
  journal={Science},
  volume={171},
  number={3972},
  pages={701--703},
  year={1971},
  publisher={American Association for the Advancement of Science}
}

@article{pylyshyn1988tracking,
  title={{Tracking multiple independent targets: Evidence for a parallel tracking mechanism}},
  author={Pylyshyn, Zenon W and Storm, Ron W},
  journal={Spatial Vision},
  volume={3},
  number={3},
  pages={179--197},
  year={1988}
}

@article{navon1977forest,
  title={{Forest before trees: The precedence of global features in visual perception}},
  author={Navon, David},
  journal={Cognitive Psychology},
  volume={9},
  number={3},
  pages={353--383},
  year={1977},
  publisher={Elsevier}
}

@article{owen2005n,
  title={{N-back working memory paradigm: A meta-analysis of normative functional neuroimaging studies}},
  author={Owen, Adrian M and McMillan, Kathryn M and Laird, Angela R and Bullmore, Ed},
  journal={Human Brain Mapping},
  volume={25},
  number={1},
  pages={46--59},
  year={2005},
  publisher={Wiley Online Library}
}

@article{cheyette2020unified,
  title={A unified account of numerosity perception},
  author={Cheyette, Samuel J and Piantadosi, Steven T},
  journal={Nature Human Behaviour},
  volume={4},
  number={12},
  pages={1265--1272},
  year={2020},
  publisher={Nature Publishing Group UK London}
}

@article{posner2016orienting,
  title={Orienting of attention: Then and now},
  author={Posner, Michael I},
  journal={The Quarterly Journal of Experimental Psychology},
  volume={69},
  number={10},
  pages={1864--1875},
  year={2016},
  publisher={Taylor \& Francis}
}

@article{clark2006reflection,
  title={{Reflection Impulsivity in Current and Former Substance Users}},
  author={Clark, Luke and Robbins, Trevor W and Ersche, Karen D and Sahakian, Barbara J},
  journal={Biological Psychiatry},
  volume={60},
  number={5},
  pages={515--522},
  year={2006},
  publisher={Elsevier}
}

@article{repp2005sensorimotor,
  title={{Sensorimotor synchronization: A review of the tapping literature}},
  author={Repp, Bruno H},
  journal={Psychonomic Bulletin \& Review},
  volume={12},
  number={6},
  pages={969--992},
  year={2005},
  publisher={Springer}
}

@article{battaglia2013simulation,
  title={Simulation as an engine of physical scene understanding},
  author={Battaglia, Peter W and Hamrick, Jessica B and Tenenbaum, Joshua B},
  journal={Proceedings of the National Academy of Sciences},
  volume={110},
  number={45},
  pages={18327--18332},
  year={2013},
  publisher={National Academy of Sciences}
}

@article{jazayeri2010temporal,
  title={Temporal context calibrates interval timing},
  author={Jazayeri, Mehrdad and Shadlen, Michael N},
  journal={Nature Neuroscience},
  volume={13},
  number={8},
  pages={1020--1026},
  year={2010},
  publisher={Nature Publishing Group US New York}
}

@article{rajsic2014asymmetrical,
  title={Asymmetrical access to color and location in visual working memory},
  author={Rajsic, Jason and Wilson, Daryl E},
  journal={Attention, Perception, \& Psychophysics},
  volume={76},
  number={7},
  pages={1902--1913},
  year={2014},
  publisher={Springer}
}

@article{dehaene1991wisconsin,
  title={{The Wisconsin Card Sorting Test: Theoretical Analysis and Modeling in a Neuronal Network}},
  author={Dehaene, Stanislas and Changeux, Jean-Pierre},
  journal={Cerebral Cortex},
  volume={1},
  number={1},
  pages={62--79},
  year={1991},
  publisher={Oxford University Press}
}

@article{eckstein2011visual,
  title={{Visual search: A retrospective}},
  author={Eckstein, Miguel P},
  journal={Journal of Vision},
  volume={11},
  number={5},
  pages={14--14},
  year={2011},
  publisher={The Association for Research in Vision and Ophthalmology}
}

@article{nowak2000fairness,
  title={{Fairness Versus Reason in the Ultimatum Game}},
  author={Nowak, Martin A and Page, Karen M and Sigmund, Karl},
  journal={Science},
  volume={289},
  number={5485},
  pages={1773--1775},
  year={2000},
  publisher={American Association for the Advancement of Science}
}

@article{maxwell2025predictive,
  title={{Predictive alternating runs and random task-switching sequences produce dissociative switch costs in the Consonant–Vowel/Odd–Even task}},
  author={Maxwell, Nicholas P and Huff, Mark J and Namias, Jacob M},
  journal={Cognitive Processing},
  volume={26},
  number={1},
  pages={157--170},
  year={2025},
  publisher={Springer}
}

@article{kunzell2017validation,
  title={{Validation of the Continuous Tracking Paradigm for Studying Implicit Motor Learning}},
  author={K{\"u}nzell, Stefan and Sie{\ss}meir, Dominicus and Ewolds, Harald},
  journal={Experimental Psychology},
  year={2017},
  publisher={Hogrefe Publishing}
}

@inproceedings{regan2014human,
  title={{Human and Computer Preferences at Chess.}},
  author={Regan, Kenneth Wingate and Biswas, Tamal and Zhou, Jason},
  booktitle={MPREF@AAAI},
  year={2014}
}

@article{stroop1935studies,
  title={Studies of interference in serial verbal reactions.},
  author={Stroop, J Ridley},
  journal={Journal of experimental psychology},
  volume={18},
  number={6},
  pages={643},
  year={1935},
  publisher={Psychological Review Company}
}
\bibliographystyle{iclr2027_conference}

\appendix
\section{Appendix}

\subsection{Task Battery Catalog}

\begin{small}
\setlength{\tabcolsep}{3pt}
\begin{longtable}{
    >{\raggedright\arraybackslash}p{0.14\textwidth}
    >{\raggedright\arraybackslash}p{0.27\textwidth}
    >{\raggedright\arraybackslash}p{0.17\textwidth}
    >{\raggedright\arraybackslash}p{0.34\textwidth}
}

\caption{Full task battery process-feature catalog with task descriptions. CAPTCHA image task is split into two rows for Classic and Cross-Tile variants.}
\label{tab:CogCAPTCHA30_feature_catalog} \\
\toprule
\textbf{Task} & \textbf{Task description} & \textbf{Task source} & \textbf{Task-specific features} \\
\midrule
\endfirsthead

\toprule
\textbf{Task} & \textbf{Task Description} & \textbf{Task Source} & \textbf{Task-Specific Features} \\

\midrule
\endhead

\bottomrule
\endfoot

Attentional blink &
Rapid serial visual presentation; detect two targets with varying temporal lag &
\cite{shapiro1997attentional} &
\texttt{blink\_magnitude}, \texttt{t2\_short\_lag\_acc}, \texttt{t2\_long\_lag\_acc}, \texttt{t2\_given\_t1\_acc}, \texttt{conditional\_blink} \\

Biological motion &
Point-light walker animation; judge walking direction &
\cite{troje2002decomposing} &
\texttt{direction\_bias}, \texttt{acc\_left}, \texttt{acc\_right}, \texttt{acc\_variability} \\

Classic CAPTCHA &
Select all tiles matching a target object &
\cite{von2003captcha} &
\texttt{overselection\_ratio}, \texttt{sequential\_score}, \texttt{direction\_change\_rate} \\

Cross-Tile &
Objects span tile boundaries, requiring spatial grouping &
\cite{von2003captcha} &
\texttt{overselection\_ratio}, \texttt{sequential\_score}, \texttt{direction\_change\_rate} \\

Change detection &
Detect changes between alternating images &
\cite{luck1997capacity} &
\texttt{mean\_flicker\_cycles}, \texttt{flicker\_cv} \\

Corsi &
Spatial span task; reproduce sequence of block locations &
\cite{corsi1972human} &
\texttt{acc\_slope\_span}, \texttt{transposition\_rate}, \texttt{intrusion\_rate}, \texttt{serial\_first\_acc}, \texttt{serial\_middle\_acc}, \texttt{serial\_last\_acc}, \texttt{primacy\_recency} \\

Delay discounting &
Choose between smaller immediate vs larger delayed rewards &
\cite{mazur2013adjusting} &
\texttt{discount\_slope}, \texttt{choice\_consistency}, \texttt{log\_discount\_k} \\

Digit span &
Verbal working memory; recall digit sequences in order &
\cite{wechsler1955wechsler} &
\texttt{acc\_slope\_span}, \texttt{transposition\_rate}, \texttt{intrusion\_rate}, \texttt{serial\_first\_acc}, \texttt{serial\_middle\_acc}, \texttt{serial\_last\_acc}, \texttt{primacy\_recency} \\

Fitts &
Pointing task; click targets of varying size and distance &
\cite{fitts1954information} &
\texttt{fitts\_slope}, \texttt{fitts\_intercept}, \texttt{click\_precision\_mean}, \texttt{click\_precision\_std}, \texttt{click\_distance\_px\_mean}, \texttt{click\_distance\_px\_std} \\

Iowa Gambling Task &
Sequential card choices from decks with different reward-risk profiles &
\cite{bechara2001neurobiology} &
\texttt{early\_exploration}, \texttt{learning\_slope}, \texttt{win\_stay}, \texttt{lose\_shift}, \texttt{stickiness}, \texttt{deck\_entropy} \\

Line bisection &
Mark midpoint of a line (spatial bias) &
\cite{jewell2000pseudoneglect} &
\texttt{mean\_abs\_error}, \texttt{error\_slope\_length} \\

Mental rotation &
Judge whether rotated shapes are same or mirrored &
\cite{shepard1971mental} &
\texttt{rotation\_slope}, \texttt{angle\_acc\_delta}, \texttt{acc\_angle\_small}, \texttt{acc\_angle\_large}, \texttt{mirror\_acc\_effect} \\

MOT &
Track moving targets among distractors &
\cite{pylyshyn1988tracking} &
\texttt{load\_slope} \\

Mouselab MDP &
Explore states via mouse movements to maximize reward &
\cite{callaway2017mouselab} &
\texttt{edge\_fraction}, \texttt{mean\_dist\_to\_edge}, \texttt{turn\_rate}, \texttt{edge\_hovers\_mean}, \texttt{state\_hovers\_mean}, \texttt{unique\_edges\_explored}, \texttt{exploration\_ratio}, \texttt{pre\_move\_hovers} \\

Navon &
Global-local letters; identify large vs small letter under congruency manipulations &
\cite{navon1977forest} &
\texttt{navon\_congruency\_effect}, \texttt{global\_advantage}, \texttt{navon\_global\_cong\_acc}, \texttt{navon\_global\_incong\_acc}, \texttt{navon\_local\_cong\_acc}, \texttt{navon\_local\_incong\_acc}, \texttt{navon\_global\_congruency\_acc}, \texttt{navon\_local\_congruency\_acc}, \texttt{global\_acc\_advantage} \\

N-back &
Working memory task; detect repeats from N trials back &
\cite{owen2005n} &
\texttt{criterion}, \texttt{hit\_rate}, \texttt{fa\_rate} \\

Numerosity &
Estimate which of two sets has more items &
\cite{cheyette2020unified} &
\texttt{difficulty\_effect}, \texttt{response\_bias}, \texttt{distance\_acc\_slope} \\

Posner &
Cueing task; respond to targets with valid/invalid spatial cues &
\cite{posner2016orienting} &
\texttt{validity\_acc\_effect}, \texttt{valid\_acc}, \texttt{invalid\_acc} \\

Target pursuit &
Track moving target with cursor; continuous control task &
\cite{kunzell2017validation} &
\texttt{cursor\_smoothness}, \texttt{directional\_accuracy}, \texttt{correction\_rate}, \texttt{anticipation\_lag\_ms} \\

Rhythm tapping &
Tap in time with rhythm; measures timing variability and drift &
\cite{repp2005sensorimotor} &
\texttt{tap\_error\_mean}, \texttt{tap\_error\_std}, \texttt{tap\_drift}, \texttt{interval\_cv} \\

Sampling &
Sequential information sampling before making a decision &
\cite{clark2006reflection} &
\texttt{mean\_total\_samples}, \texttt{var\_total\_samples}, \texttt{mean\_sample\_bias}, \texttt{samples\_easy}, \texttt{samples\_medium}, \texttt{samples\_hard}, \texttt{neutral\_first\_rate}, \texttt{effort\_accuracy\_corr} \\

Sequence prediction &
Predict next element in a sequence with varying complexity &
[N/A] source &
\texttt{difficulty\_slope}, \texttt{seq\_acc\_d1}, \texttt{seq\_acc\_d2}, \texttt{seq\_acc\_d3}, \texttt{seq\_acc\_d4}, \texttt{seq\_acc\_d5} \\

Slingshot &
Aim and launch projectile to hit targets; physics-based control task &
\cite{battaglia2013simulation} &
\texttt{mean\_aim\_time}, \texttt{aim\_time\_cv}, \texttt{distance\_calibration\_r}, \texttt{mean\_angle\_delta\_45}, \texttt{angle\_variability}, \texttt{undershoot\_rate}, \texttt{overshoot\_rate} \\

Stroop &
Name ink color of words; congruent vs incongruent trials &
\cite{stroop1935studies} &
\texttt{post\_error\_slowing}, \texttt{congruency\_acc\_effect},  \texttt{cong\_acc}, \texttt{incong\_acc} \\

Task switching &
Alternate between task rules; measures switching cost &
\cite{maxwell2025predictive} &
\texttt{switch\_cost\_acc},   \texttt{repeat\_acc}, \texttt{switch\_acc} \\

Temporal reproduction &
Reproduce time intervals; measures temporal bias and variability &
\cite{jazayeri2010temporal} &
\texttt{temporal\_bias}, \texttt{mean\_ratio}, \texttt{regression\_to\_mean} \\

Tower physics &
Predict stability of stacked objects (intuitive physics) &
\cite{battaglia2013simulation} &
\texttt{complexity\_slope} \\

Ultimatum game &
Fairness decision-making &
\cite{nowak2000fairness} &
\texttt{responder\_accept\_rate}, \texttt{reject\_unfair\_rate} \\

Visual search &
Find target among distractors &
\cite{eckstein2011visual} &
\texttt{search\_slope}, \texttt{search\_click\_precision}, \texttt{click\_dist\_setsize\_slope} \\

Visual Working Memory &
Recall location/color under varying set sizes &
\cite{rajsic2014asymmetrical} & 
\texttt{error\_slope\_setsize}, \texttt{bias\_x\_mean}, \texttt{bias\_y\_mean}, \texttt{bias\_radius\_mean}, \texttt{bias\_anisotropy}, \texttt{large\_error\_frac}, \texttt{swap\_error\_rate} \\

WCST &
Infer and update sorting rule from feedback &
\cite{dehaene1991wisconsin} & 
\texttt{learning\_slope}, \texttt{post\_error\_slowing}, \texttt{shift\_error\_rate}, \texttt{perseveration\_cost}, \texttt{rule\_acc\_variability} \\

\end{longtable}
\end{small}

\subsection{Full task results}

\begin{table}[ht]
\centering
\caption{Performance across combinations of observed and left-out features. The results show that the reduction in the human-agent process gap is attenuated when evaluation features are not observed during training.}

\label{tab:feature_combinations}
\begin{tabular}{clcc}
\toprule
\textbf{Combination} & \textbf{Feature set} & \textbf{P-SFT} & \textbf{A-SFT} \\
\midrule
\multirow{2}{*}{1}
    & Observed features & 0.06  & 0.62 \\
    & Left-out features & 0.32  & 0.66 \\
\midrule
\multirow{2}{*}{2}
    & Observed features & 0.13  & 0.40 \\
    & Left-out features & 0.35  & 0.44 \\
\midrule
\multirow{2}{*}{3}
    & Observed features & 0.08  & 0.47 \\
    & Left-out features & 0.321 & 0.55 \\
\bottomrule
\end{tabular}
\end{table}

\begin{table}[ht]
\centering
\caption{
Distributional distance to humans measured by mean absolute Cohen's $d$ across process features (lower is more human-like). Centaur more closely matched human process distributions than frontier agents across all evaluated tasks.
}
\label{tab:cross_task}

\begin{tabular}{lcccc}
\toprule
Method & Sampling & IGT & WCST & Average \\
\midrule
Base Qwen (1.5B) & 0.89 & 0.81 & 0.88 & 0.86 \\
Claude Sonnet & 1.15 & 3.32 & 0.48 & 1.65 \\
GPT-5 & 0.27 & 1.91 & 1.91 & 1.36 \\
Gemini 2.5 Pro & 1.26 & 0.56 & 1.38 & 1.07 \\
Centaur (70B) & 0.15 & 0.42 & 0.76 & 0.44 \\
A-SFT (1.5B) & 0.51 & 0.42 & 0.93 & 0.62 \\
A-SFT (7B) & 0.17 & 0.18 & 0.46 & 0.27 \\
A-SFT (14B) & 0.27 & 0.23 & 0.62 & 0.38 \\
A-SFT (32B) & 0.47 & 0.11 & 0.15 & 0.25 \\
A-SFT (70B) & 0.10 & 0.24 & 1.12 & 0.48 \\
\textbf{P-SFT (1.5B)} & \textbf{0.11} & \textbf{0.06} & \textbf{0.03} & \textbf{0.06} \\
P-SFT (7B) & 0.07 & 0.14 & 0.30 & 0.17 \\
P-SFT (14B) & 0.30 & 0.22 & 0.18 & 0.23 \\
P-SFT (32B) & 0.37 & 0.11 & 0.23 & 0.23 \\
P-SFT (70B) & 0.76 & 0.13 & 0.31 & 0.40 \\
\midrule
\multicolumn{5}{l}{\emph{Left-out features}} \\
Centaur (70B) & 0.25 & 0.48 & 0.48 & 0.40 \\
A-SFT (1.5B) & 0.68 & 0.41 & 0.90 & 0.66 \\
A-SFT (7B) & 0.17 & 0.08 & 0.46 & 0.23 \\
A-SFT (14B) & 0.34 & 0.32 & 0.73 & 0.46 \\
A-SFT (32B) & 0.36 & 0.07 & 0.10 & 0.18 \\
A-SFT (70B) & 0.17 & 0.17 & 0.73 & 0.36 \\
P-SFT (1.5B) & 0.39 & 0.36 & 0.21 & 0.32 \\
P-SFT (7B) & 0.51 & 0.55 & 0.50 & 0.52 \\
P-SFT (14B) & 0.35 & 0.21 & 0.16 & 0.24 \\
P-SFT (32B) & 0.38 & 0.09 & 0.03 & 0.17 \\
P-SFT (70B) & 0.78 & 0.51 & 0.38 & 0.55 \\
\midrule
\multicolumn{5}{l}{\emph{Cross-task}} \\
A-SFT (1.5B) & 0.74 & 0.31 & 0.84 & 0.63 \\
A-SFT (7B) & 0.29 & 0.33 & 0.83 & 0.49 \\
A-SFT (14B) & 0.69 & 0.09 & 1.08 & 0.62 \\
A-SFT (32B) & 0.61 & 0.29 & 0.92 & 0.61 \\
A-SFT (70B) & 0.59 & 0.34 & 1.34 & 0.76 \\
P-SFT (1.5B) & 1.00 & 0.46 & 0.73 & 0.73 \\
P-SFT (7B) & 0.28 & 0.36 & 0.89 & 0.51 \\
P-SFT (14B) & 0.76 & 0.16 & 1.21 & 0.71 \\
P-SFT (32B) & 0.61 & 0.28 & 0.92 & 0.60 \\
P-SFT (70B) & 0.54 & 0.28 & 1.37 & 0.73 \\
\bottomrule
\end{tabular}

\end{table}

\begin{table}[ht]
\centering
\caption{Combined classifier fool rate. A Random Forest classifier (AUC = 0.89 $\pm$ 0.03) was trained to distinguish humans from off-the-shelf LLM agents using process features concatenated across all three tasks. Test set off-the-shelf runs, fine-tuned models and Centaur were held out from training and evaluated against the fitted classifier. Higher fool rate indicates more human-like behavior.}\label{tab:fool_combined}

\begin{tabular}{lcc}
\toprule
Method & Fool rate $\uparrow$ & Mean $P$(human) \\
\midrule
\multicolumn{3}{l}{\emph{Off-the-shelf models (held-out evaluation runs)}} \\
GPT-5 & 7.6\% & 0.18 \\
Gemini 2.5 Pro & 0.0\% & 0.16 \\
Claude Sonnet & 0.0\% & 0.05 \\
Base Qwen (1.5B) & 6.8\% & 0.26 \\
Base Qwen (7B) & 17.4\% & 0.34 \\
Base Qwen (14B) & 8.6\% & 0.27 \\
Base Qwen (32B) & 3.4\% & 0.26 \\
Base Llama (70B) & 18.4\% & 0.33 \\
\midrule
Centaur (70B) & 44.0\% & 0.45 \\
A-SFT (1.5B) & 27.5\% & 0.37 \\
A-SFT (7B) & 52.5\% & 0.50 \\
A-SFT (14B) & 38.5\% & 0.43 \\
A-SFT (32B) & 47.5\% & 0.50 \\
A-SFT (70B) & 45.5\% & 0.48 \\
P-SFT (1.5B) & 74.0\% & 0.64 \\
P-SFT (7B) & 68.0\% & 0.62 \\
P-SFT (14B) & 52.5\% & 0.52 \\
P-SFT (32B) & 45.0\% & 0.47 \\
\textbf{P-SFT (70B)} & \textbf{90.5\%} & \textbf{0.73} \\
\midrule
\multicolumn{3}{l}{\emph{Left-out features}} \\
Centaur (70B) & 43.0\% & 0.48 \\
A-SFT (1.5B) & 13.0\% & 0.30 \\
A-SFT (7B) & 48.0\% & 0.50 \\
A-SFT (14B) & 25.0\% & 0.36 \\
A-SFT (32B) & 33.0\% & 0.44 \\
A-SFT (70B) & 52.0\% & 0.54 \\
P-SFT (1.5B) & 40.0\% & 0.46 \\
P-SFT (7B) & 22.0\% & 0.36 \\
P-SFT (14B) & 46.0\% & 0.45 \\
P-SFT (32B) & 29.0\% & 0.40 \\
P-SFT (70B) & 59.0\% & 0.54 \\
\midrule
\multicolumn{3}{l}{\emph{Cross-task}} \\
A-SFT (1.5B) & 20.5\% & 0.33 \\
A-SFT (7B) & 26.0\% & 0.36 \\
A-SFT (14B) & 11.5\% & 0.27 \\
A-SFT (32B) & 11.0\% & 0.30 \\
A-SFT (70B) & 16.0\% & 0.31 \\
P-SFT (1.5B) & 31.0\% & 0.39 \\
P-SFT (7B) & 21.0\% & 0.33 \\
P-SFT (14B) & 10.0\% & 0.26 \\
P-SFT (32B) & 18.6\% & 0.33 \\
P-SFT (70B) & 18.0\% & 0.34 \\
\bottomrule
\end{tabular}

\end{table}

\begin{table*}[ht]
\centering
\small

\caption{
Energy distance---a multivariate distributional metric defined as
$2\,\mathbb{E}\|X-Y\|-\mathbb{E}\|X-X'\|-\mathbb{E}\|Y-Y'\|$,
computed on per-feature human-standardized features---captures both
location and shape differences between a method's and humans' joint
feature distributions; a value of zero indicates identical distributions.
}

\label{tab:energy}
\begin{tabular}{lcccc}
\toprule
\textbf{Method} & \textbf{Sampling} & \textbf{IGT} &
\textbf{WCST} & \textbf{Average} \\
\midrule

Base Qwen (1.5B) & 1.50 & 1.27 & 1.04 & 1.27 \\
Claude Sonnet    & 1.33 & 11.14 & 0.61 & 4.36 \\
GPT-5            & 0.29 & 5.84 & 2.28 & 2.81 \\
Gemini 2.5 Pro   & 2.06 & 1.30 & 1.42 & 1.59 \\

\midrule
\multicolumn{5}{l}{\textit{Observed features}} \\
\addlinespace[2pt]

Centaur (70B)       & 0.03 & 0.52 & 0.56 & 0.37 \\
A-SFT (1.5B)        & 0.55 & 0.53 & 0.97 & 0.68 \\
A-SFT (7B)          & 0.07 & 0.16 & 0.21 & 0.15 \\
A-SFT (14B)         & 0.16 & 0.16 & 0.47 & 0.26 \\
A-SFT (32B)         & 0.42 & 0.10 & 0.04 & 0.19 \\
A-SFT (Llama 70B)   & 0.07 & 0.14 & 1.04 & 0.42 \\
\textbf{P-SFT (1.5B)}
                    & \textbf{0.11} & \textbf{0.06}
                    & \textbf{0.06} & \textbf{0.07} \\
P-SFT (7B)          & 0.09 & 0.16 & 0.13 & 0.13 \\
P-SFT (14B)         & 0.16 & 0.14 & 0.10 & 0.14 \\
P-SFT (32B)         & 0.23 & 0.05 & 0.07 & 0.12 \\
P-SFT (Llama 70B)
                    & 0.48 & 0.15 & 0.10 & 0.24 \\

\midrule
\multicolumn{5}{l}{\textit{Left-out features}} \\
\addlinespace[2pt]

Centaur (70B)       & 0.15 & 0.29 & 0.16 & 0.20 \\
A-SFT (1.5B)        & 1.24 & 0.26 & 0.60 & 0.70 \\
A-SFT (7B)          & 0.12 & 0.03 & 0.20 & 0.11 \\
A-SFT (14B)         & 0.45 & 0.13 & 0.44 & 0.34 \\
A-SFT (32B)         & 0.45 & 0.03 & 0.02 & 0.17 \\
A-SFT (70B)         & 0.14 & 0.02 & 0.40 & 0.18 \\
P-SFT (1.5B)        & 0.41 & 0.33 & 0.04 & 0.26 \\
P-SFT (7B)          & 1.50 & 0.71 & 0.24 & 0.81 \\
P-SFT (14B)         & 0.32 & 0.07 & 0.03 & 0.14 \\
P-SFT (32B)         & 0.45 & 0.04 & 0.01 & 0.17 \\
P-SFT (70B)
                    & 1.15 & 0.57 & 0.09 & 0.60 \\

\midrule
\multicolumn{5}{l}{\textit{Cross-task}} \\
\addlinespace[2pt]

A-SFT (1.5B)        & 1.05 & 0.39 & 0.83 & 0.76 \\
A-SFT (7B)          & 0.23 & 0.38 & 0.73 & 0.45 \\
A-SFT (14B)         & 0.78 & 0.08 & 1.13 & 0.66 \\
A-SFT (32B)         & 0.61 & 0.24 & 0.89 & 0.58 \\
A-SFT (70B)         & 0.54 & 0.29 & 1.54 & 0.79 \\
P-SFT (1.5B)        & 1.28 & 0.44 & 0.59 & 0.77 \\
P-SFT (7B)          & 0.24 & 0.42 & 0.95 & 0.54 \\
P-SFT (14B)         & 0.78 & 0.21 & 1.42 & 0.80 \\
P-SFT (32B)         & 0.57 & 0.16 & 0.86 & 0.53 \\
P-SFT (70B)
                    & 0.40 & 0.14 & 1.53 & 0.69 \\

\bottomrule
\end{tabular}

\end{table*}

\begin{figure}[ht]
    \centering
    \includegraphics[width=0.90\linewidth]{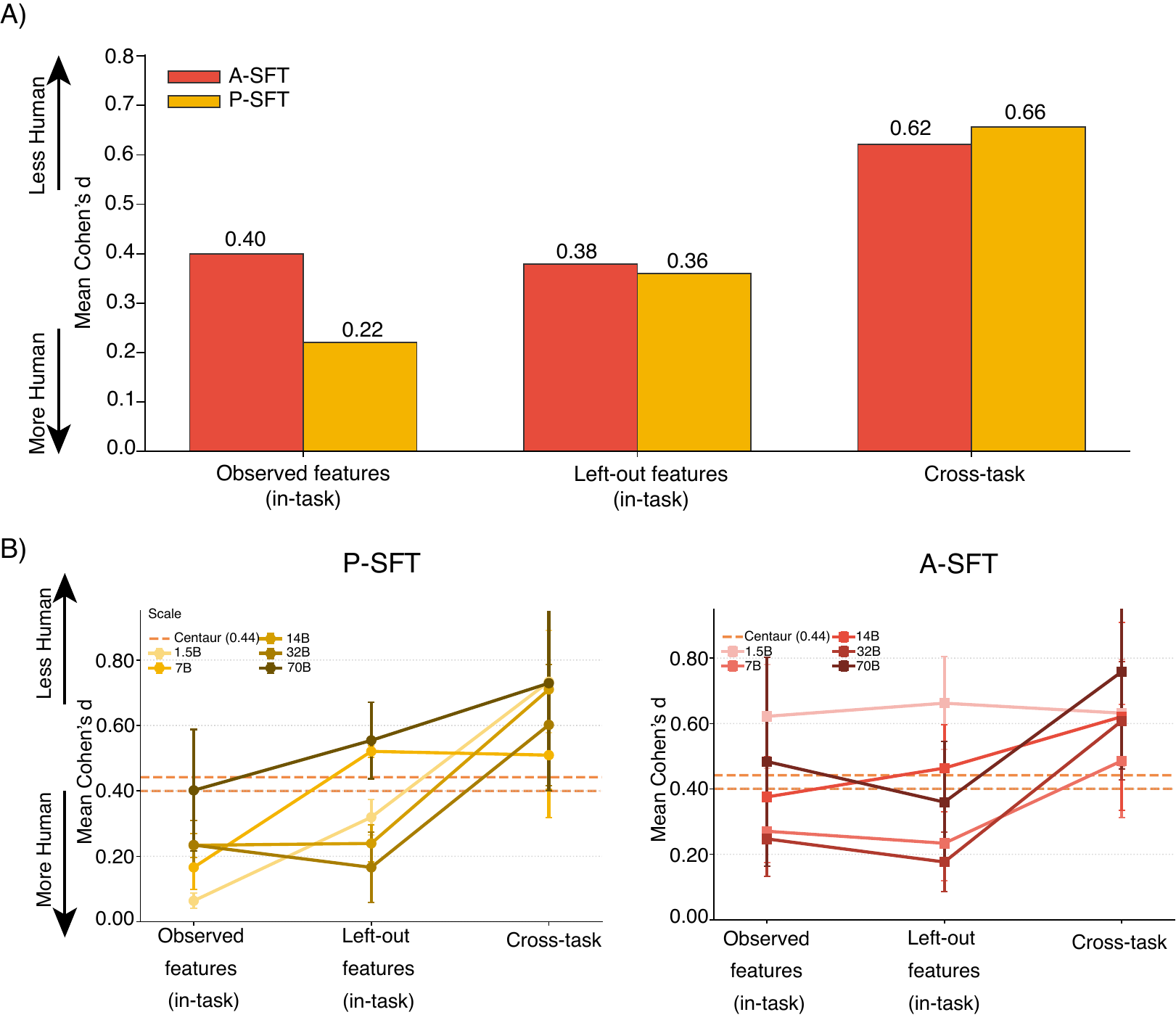}
    \caption{P-SFT models show lower overall distance to human process distributions than A-SFT models. This advantage diminishes when evaluated on left-out features from the same task and largely disappears under cross-task evaluation.}
    \label{fig:distance-all-models}
\end{figure}

\begin{figure}[ht]
    \centering
    \includegraphics[width=0.90\linewidth]{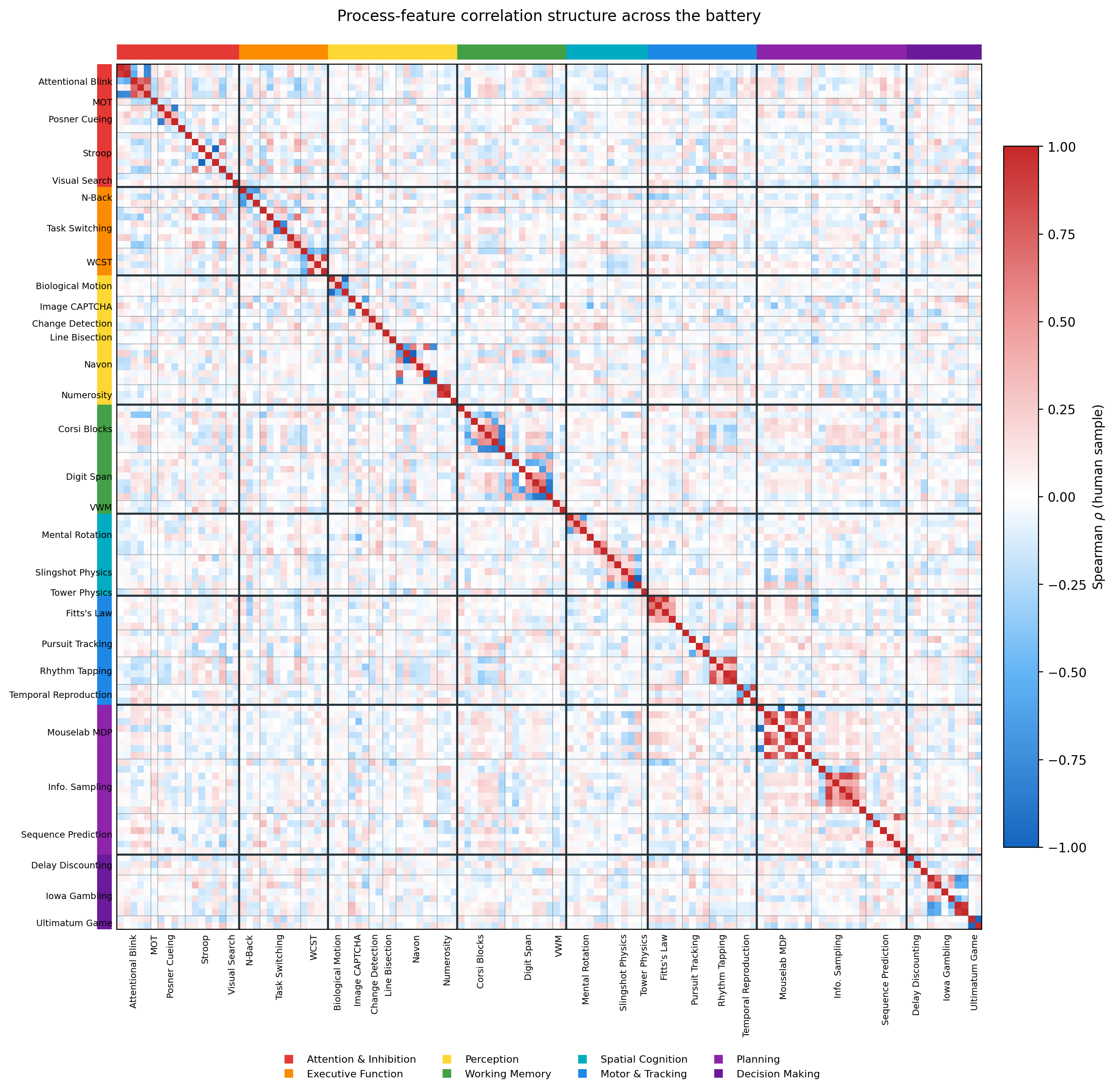}
    \caption{Correlation of features across the full task battery. Features across tasks show generally low correlations, indicating limited overlap in the measured process dimensions.}
    \label{fig:feature-correlation-matrix}
\end{figure}

\FloatBarrier

\subsection{P-SFT: implementation details}
\label{app:psft_details}

All fine-tuning was performed on a single NVIDIA H200 SXM GPU (141\,GB VRAM). We fine-tuned Qwen2.5-Instruct \citep{qwen2.5} models (1.5B, 7B, 14B, 32B) and Llama3.3-Instruct (70B) via LoRA \citep{hu2021lora} (rank 8, alpha 16, dropout 0.05) applied to all attention and MLP projection layers. Training used AdamW with learning rate $10^{-5}$, gradient clipping at norm 1, and gradient checkpointing to fit the per-rollout state expansion in memory. A-SFT trained for 300 steps of cross-entropy on human actions. P-SFT added 30 steps of cross-entropy warmup followed by up to 150 steps of joint optimization, with early stopping when the feature-matching loss plateaued (patience of 20--50 steps depending on task). At each joint step, we sampled a fresh stimulus sequence from a random training participant's run, preventing the model from overfitting to a single sequence of task stimuli.

\paragraph{Evaluation.} We use 2-fold cross-validation: 97 human participants are split into two folds (47 and 50). Each fold's model is trained on one half and evaluated against held-out humans from the other half. Per-fold rollouts (100 per fold, 200 total) are compared to per-fold held-out humans, and results are pooled. This ensures no overlap between training and evaluation participants.

\paragraph{Information Sampling.} This task involves multiple decisions per trial: the agent chooses to reveal a tile from option A, reveal a tile from option B, or stop and choose an option. The run-level features are mean and variance of the per-trial total tile count. Because each trial involves a variable-length sequence of decisions, we compute the expected tile count differentiably by expanding the set of reachable game states at each within-trial step, weighted by the policy's action probabilities. States with identical observable histories (e.g., two A-tiles revealed via different orderings) are merged. The expected stopping time is then
\begin{equation}
\mathbb{E}[T] = \sum_{t=0}^{T_{\max}} t \cdot P(\text{stop at step } t),
\end{equation}
computed exactly by accumulating $P(\text{stop} \mid \text{state}_t) \cdot P(\text{state}_t)$ over the expansion. We cap the frontier at 16 active states per step, which exceeds the theoretical maximum of 11 distinct observable states after merging (since with 5 tiles per option, the merged state space at step $t$ is the set of ($n_A$,$n_B$) pairs with $n_A+n_B =t$). At trial boundaries, we select the most-likely outcome from the frontier (greedy stop time and choice) to construct the prompt continuation for subsequent trials, matching the prompt format used at evaluation time.

\paragraph{Iowa Gambling Task.} Each trial is a single 4-way choice over decks A, B, C, D; no within-trial state expansion is needed. The six run features, all differentiable in the model's action probabilities $p_t$, are:
\begin{itemize}[nosep]
\item \emph{learning\_slope}: mean good-deck probability in the second half minus the first half, $\frac{1}{\lfloor T/2 \rfloor}\sum_{t \geq \lceil T/2 \rceil} p_t^{\text{good}} - \frac{1}{\lceil T/2 \rceil}\sum_{t < \lceil T/2 \rceil} p_t^{\text{good}}$, where $p_t^{\text{good}} = p_t(C) + p_t(D)$
\item \emph{stickiness}: $\frac{1}{T-1}\sum_{t=2}^{T} p_t(a_{t-1})$, the mean probability assigned to the previous trial's sampled action (the sampled action $a_{t-1}$ is treated as fixed; gradients flow through $p_t$ only)
\item \emph{deck\_entropy}: $H(\bar{p}) = -\sum_d \bar{p}_d \log \bar{p}_d$, where $\bar{p} = \frac{1}{T}\sum_t p_t$ is the mean action distribution across trials
\item \emph{win\_stay}: mean probability of repeating the previous action on trials following a positive outcome, $\frac{1}{|\mathcal{W}|}\sum_{t \in \mathcal{W}} p_t(a_{t-1})$, where $\mathcal{W} = \{t : \text{net}_{t-1} > 0\}$
\item \emph{lose\_shift}: mean probability of switching away from the previous action on trials following a non-positive outcome, $\frac{1}{|\mathcal{L}|}\sum_{t \in \mathcal{L}} (1 - p_t(a_{t-1}))$, where $\mathcal{L} = \{t : \text{net}_{t-1} \leq 0\}$
\item \emph{good\_deck\_rate}: $\frac{1}{T}\sum_t (p_t(C) + p_t(D))$
\end{itemize}
Win/loss classification at each trial is determined by the sampled action and the deterministic deck outcomes, and is treated as fixed for gradient computation. Gradients flow only through the probability vectors $p_t$.

\paragraph{Wisconsin Card Sorting Task.} Each trial is a single 4-way choice (match to reference card 1, 2, 3, or 4). Whether the choice is correct depends on the current hidden rule and the test card. The two run features are:
\begin{itemize}[nosep]
\item \emph{perseveration\_cost}: mean $p(\text{correct})$ on non-shift trials minus mean $p(\text{correct})$ on shift trials, where $p(\text{correct}) = p_t(\text{correct index})$ is the probability the model assigns to the correct card on trial $t$
\item \emph{learning\_slope}: mean $p(\text{correct})$ in the second half of the run minus the first half
\end{itemize}
Both are direct functions of the model's per-trial softmax probabilities evaluated at the correct action index, which is known from the stimulus sequence.

\paragraph{Training stability.} Under the closed-form feature estimator, the feature-matching loss decreases monotonically during phase-2 training across all three tasks, typically reaching a plateau around step 90-100. We use AdamW with learning rate $10^{-5}$, gradient clipping at norm 1, and gradient checkpointing to fit the per-rollout computation in memory.

\subsection{Task Instructions and Agent Prompts}
\label{app:task_prompts}

\subsubsection{Information Sampling}

\paragraph{Agent Prompt (Sampling Phase)}
\begin{agentbox}
There are two rows of tiles:

\begin{itemize}[nosep]
    \item Option A (top/blue): 5 tiles
    \item Option B (bottom/red): 5 tiles
\end{itemize}

Hidden tiles show ``?''. Revealed tiles show numbers (0--100). Your goal is to determine which option has the higher average.

Decide whether to sample another tile or stop and choose.

If sampling, respond with:
\texttt{\{"action": "sample", "option": "A", "tile": <0--4>\}}

or

\texttt{\{"action": "sample", "option": "B", "tile": <0--4>\}}

If ready to choose:
\texttt{\{"action": "choose"\}}
\end{agentbox}

\paragraph{Agent Prompt (Choice Phase)}
\begin{agentbox}
Based on the revealed tile values, which option has the higher average?

Respond with ONLY:

\texttt{\{"choice": "A"\}} or \texttt{\{"choice": "B"\}}
\end{agentbox}

\paragraph{Participant Instructions}
\begin{humanbox}
You will see two rows of 5 hidden tiles: Option A (top/blue) and Option B (bottom/red).

Each tile hides a number. One option tends to have higher values than the other.

Click a tile to reveal its value. Each flip costs points.

Choosing correctly earns bonus points.

When you are confident, click ``I'm ready to choose.''

Your goal is to determine which option has higher values overall while minimizing flip costs.
\end{humanbox}

\subsubsection{Visual Working Memory}

\paragraph{Agent Prompt}
\begin{agentbox}

Image 1 shows colored squares during encoding. Memorize each color's position.

Image 2 shows an empty canvas and a target color labeled ``Click where this color was.''

Find where the target color was located and report its center coordinates.

Respond with ONLY:

\texttt{\{"x": <number>, "y": <number>\}}
\end{agentbox}

\paragraph{Participant Instructions}
\begin{humanbox}
You will see several colored squares flash on the screen.

After they disappear, one of the colors will appear as the target.

Click where on the screen that color was previously shown.

\end{humanbox}

\subsubsection{Visual Search}

\paragraph{Agent Prompt}
\begin{agentbox}

Find the RED CIRCLE and click on it.

It will be hidden among red squares and blue circles.

Respond with ONLY:

\texttt{\{"x": <number>, "y": <number>\}}
\end{agentbox}

\paragraph{Participant Instructions}
\begin{humanbox}
Find the RED CIRCLE and click on it.

It will be hidden among red squares and blue circles.

\end{humanbox}

\subsubsection{Multiple Object Tracking}

\paragraph{Agent Prompt (Highlight Phase)}
\begin{agentbox}
Some dots are highlighted as targets.

Report the coordinates of each highlighted target.

Respond with ONLY:

\texttt{\{"targets": [\{"x": N, "y": N\}, ...]\}}
\end{agentbox}

\paragraph{Agent Prompt (Tracking Phase)}
\begin{agentbox}
Track the previously highlighted dots as they move.

Update each target's position to the nearest dot.

Respond with ONLY:

\texttt{\{"targets": [\{"x": N, "y": N\}, ...]\}}
\end{agentbox}

\paragraph{Participant Instructions}
\begin{humanbox}
Some dots will flash yellow---these are your targets.

All dots will then begin moving. Track the targets.

When the dots stop, click the dots you believe were the targets.
\end{humanbox}

\subsubsection{Delay Discounting}

\paragraph{Agent Prompt}
\begin{agentbox}

Look at the two monetary options on screen: A smaller amount now or A larger amount later

Determine the amounts and delay, reason about the tradeoff, and report your choice.

Respond as JSON:

\texttt{\{"now\_amount": N, "later\_amount": N, "delay": "X", "choice": "now" or "later"\}}
\end{agentbox}

\paragraph{Participant Instructions}
\begin{humanbox}
On each trial you will choose between:

\begin{itemize}[nosep]
    \item A smaller amount now
    \item A larger amount later
\end{itemize}

There are no right or wrong answers---choose whichever you prefer.
\end{humanbox}

\end{document}